\documentclass[10pt,twocolumn,letterpaper]{article}

\usepackage{cvpr}
\usepackage{times}
\usepackage{epsfig}
\usepackage{graphicx}
\usepackage{amsmath}
\usepackage{amssymb}
\usepackage{booktabs}
\usepackage{color}
\usepackage[table]{xcolor}

\definecolor{col1}{RGB}{234,209,220}
\definecolor{col2}{RGB}{217,234,211}
\newcommand*{\ea}{{\em et al.}\@\xspace}

\DeclareMathOperator*{\argmin}{arg\,min}

\newcommand{\refsec}[1]{Sec.~\ref{sec:#1}}
\newcommand{\reffig}[1]{Fig.~\ref{fig:#1}}
\newcommand{\refeq}[1]{Eq.~\ref{eq:#1}}
\newcommand{\reftab}[1]{Table~\ref{tab:#1}}

\renewcommand{\eg}[1][ ]{{\em e.\thinspace{}g\@.{}}#1}

\usepackage[pagebackref=true,breaklinks=true,letterpaper=true,colorlinks,bookmarks=false]{hyperref}

\cvprfinalcopy 

\def\cvprPaperID{1244} 

\ifcvprfinal\fi
\begin{document}

\title{Unsupervised Learning of Depth and Ego-Motion from Monocular Video \\Using 3D Geometric Constraints}

\author{Reza Mahjourian\\
University of Texas at Austin, Google Brain\\
\and
Martin Wicke\\
Google Brain\\
\and
Anelia Angelova\\
Google Brain\\
}

\maketitle

\begin{abstract}
We present a novel approach for unsupervised learning of depth and ego-motion from monocular video. Unsupervised learning removes the need for separate supervisory signals (depth or ego-motion ground truth, or multi-view video).  Prior work in unsupervised depth learning uses pixel-wise or gradient-based losses, which only consider pixels in small local neighborhoods. Our main contribution is to explicitly consider the inferred 3D geometry of the whole scene, and enforce consistency of the estimated 3D point clouds and ego-motion across consecutive frames. This is a challenging task and is solved by a novel (approximate) backpropagation algorithm for aligning 3D structures. 

We combine this novel 3D-based loss with 2D losses based on photometric quality of frame reconstructions using estimated depth and ego-motion from adjacent frames.  We also incorporate validity masks to avoid penalizing areas in which no useful information exists.

We test our algorithm on the KITTI dataset and on a video dataset captured on an uncalibrated mobile phone camera. Our proposed approach consistently improves depth estimates on both datasets, and outperforms the state-of-the-art for both depth and ego-motion.  Because we only require a simple video, learning depth and ego-motion on large and varied datasets becomes possible.  We demonstrate this by training on the low quality uncalibrated video dataset and evaluating on KITTI, ranking among top performing prior methods which are trained on KITTI itself.~\footnote{Code and data available at \href{http://sites.google.com/view/vid2depth}{http://sites.google.com/view/vid2depth}}
\end{abstract}

\begin{figure}[t]
  \begin{center}
	  \includegraphics[width=1.0\linewidth]{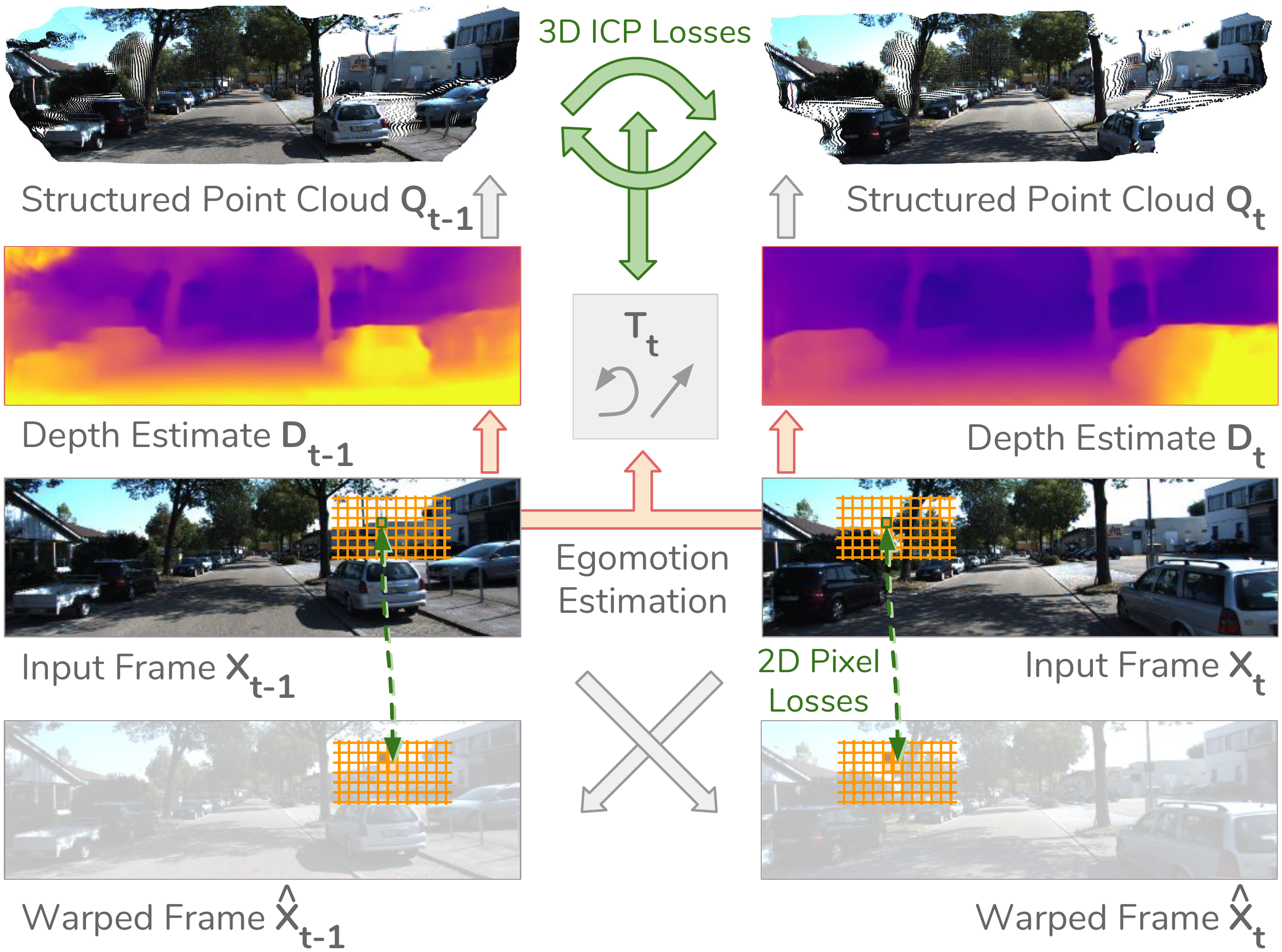}
  \end{center}
  \caption{Overview of our method. In addition to 2D photometric losses, novel 3D geometric losses are used as supervision to adjust unsupervised depth and ego-motion estimates by the neural network.  Orange arrows represent model's predictions.  Gray arrows represent mechanical transformations.  Green arrows represent losses.  The depth images shown are sample outputs from our trained model.}
  \label{fig:approach}
  \vspace{-10pt}
\end{figure}

\section{Introduction}

Inferring the depth of a scene and one's ego-motion is one of the key challenges in fields such as robotics and autonomous driving. Being able to estimate the exact position of objects in 3D and the scene geometry is essential for motion planning and decision making.

Most supervised methods for learning depth and ego-motion require carefully calibrated setups.  This severely limits the amount and variety of training data they can use, which is why supervised techniques are often applied only to a number of well-known datasets like KITTI~\cite{geiger2013vision} and Cityscapes~\cite{cordts2016cityscapes}.  Even when ground-truth depth data is available, it is often imperfect and causes distinct prediction artifacts.  Rotating LIDAR scanners cannot produce depth that temporally aligns with the corresponding image taken by a camera---even if the camera and LIDAR are carefully synchronized.  Structured light depth sensors---and to a lesser extent, LIDAR and time-of-flight sensors---suffer from noise and structural artifacts, especially in presence of reflective, transparent, or dark surfaces.  Lastly, there is usually an offset between the depth sensor and the camera, which causes gaps or overlaps when the point cloud is projected onto the camera's viewpoint.  These problems lead to artifacts in models trained on such data.

This paper proposes a method for unsupervised learning of depth and ego-motion from monocular (single-camera) videos. The only form of supervision that we use comes from assumptions about consistency and temporal coherence between consecutive frames in a monocular video (camera intrinsics are also used).

Cameras are by far the best understood and most ubiquitous sensor available to us.  High quality cameras are inexpensive and easy to deploy.  The ability to train on arbitrary monocular video opens up virtually infinite amounts of training data, without sensing artifacts or inter-sensor calibration issues. 

In order to learn depth in a completely unsupervised fashion, we rely on existence of ego-motion in the video.  Given two consecutive frames from the video, a neural network produces single-view depth estimates from each frame, and an ego-motion estimate from the frame pair.  Requiring that the depth and ego-motion estimates from adjacent frames are consistent serves as supervision for training the model.  This method allows learning depth because the transformation from depth and ego-motion to a new frame is well understood and a good approximation can be written down as a fixed differentiable function.

\begin{figure}[t]
  \begin{center}
	  \includegraphics[width=0.9\linewidth]{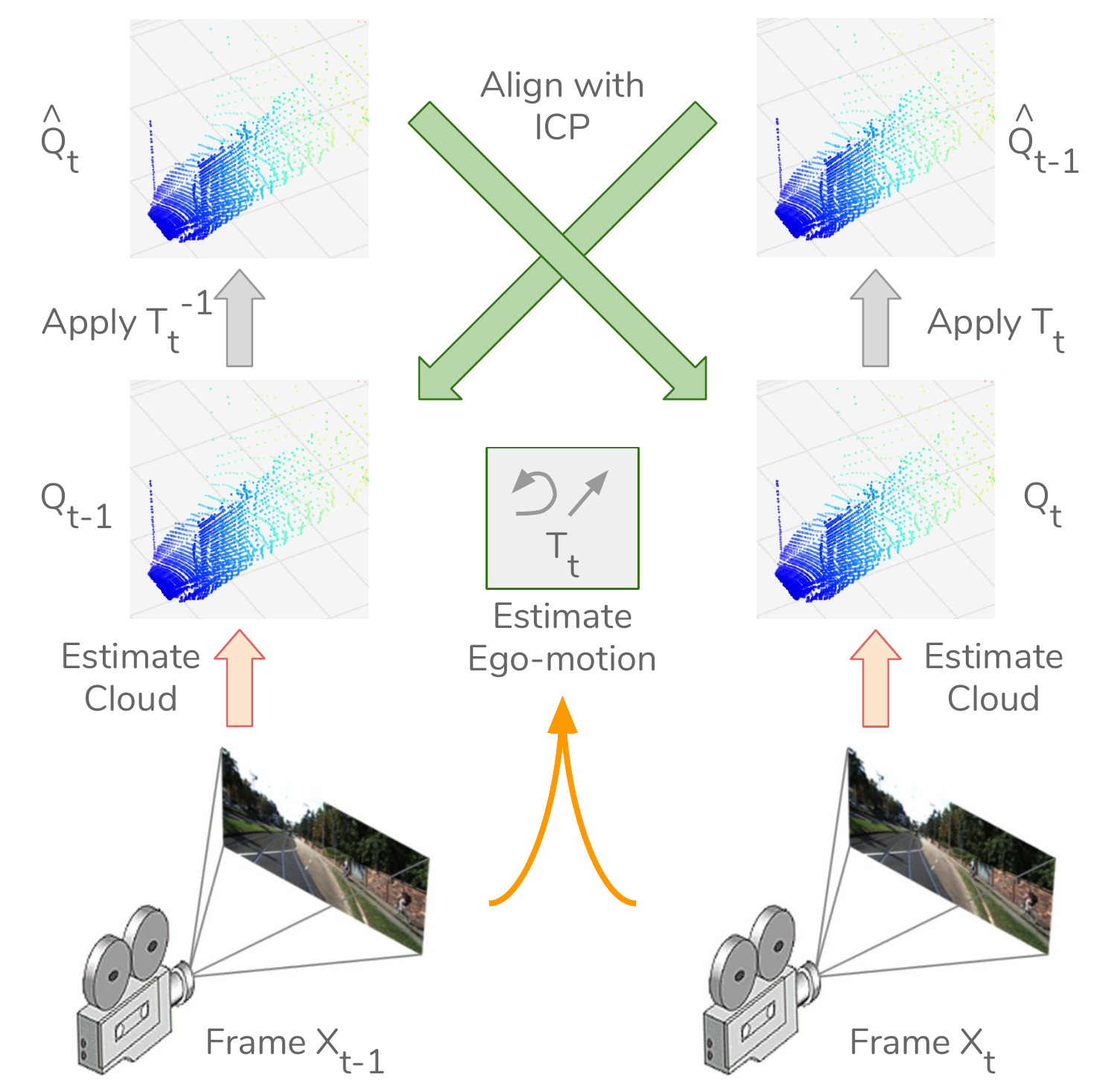}
  \end{center}
  \caption{The 3D loss: ICP is applied symmetrically in forward and backward directions to bring the depth and ego-motion estimates from two consecutive frames into agreement.  The products of ICP generate gradients which are used to improve the depth and ego-motion estimates.}
  \label{fig:icp}
\end{figure}

Our main contributions are the following:

{\bf Imposing 3D constraints.} We propose a loss which directly penalizes inconsistencies in the estimated depth without relying on backpropagation through the image reconstruction process. We compare depth extracted from adjacent frames by directly comparing 3D point clouds in a common reference frame.  Intuitively, assuming there is no significant object motion in the scene, one can transform the estimated point cloud for each frame into the predicted point cloud for the other frame by applying ego-motion or its inverse (\reffig{approach} and \reffig{icp}).

To the best of our knowledge, our approach is the first depth-from-video algorithm to use 3D information in a differentiable loss function. Our experiments show that adding losses computed directly on the 3D geometry improves results significantly.

{\bf Principled masking.}
When transforming a frame and projecting it, some parts of the scene are not covered in the new view (either due to parallax effects or because objects left or entered the field of view). Depth and image pixels in those areas are not useful for learning; using their values in the loss degrades results. Previous methods have approached this problem by adding a general-purpose learned mask to the model~\cite{zhou2017unsupervised}, or applying post-processing to remove edge artifacts~\cite{godard2017monodepth}.  However, learning the mask is not very effective.  Computing the masks analytically leaves a simpler learning problem for the model.

{\bf Learning from an uncalibrated video stream.} We demonstrate that our proposed approach can consume and learn from any monocular video source with camera motion.  We record a new dataset containing monocular video captured using a hand-held commercial phone camera while riding a bicycle.  We train our depth and ego-motion model only on these videos, then evaluate the quality its predictions by testing the trained model on the KITTI dataset.

\section{Related Work}
\label{sec:related-work}

Classical solutions to depth and ego-motion estimation involve stereo and feature matching techniques~\cite{scharstein2002taxonomy}, whereas recent methods have shown success using deep learning~\cite{flynn2016deepstereo}.

Most pioneering works that learn depth from images rely on supervision from depth sensors~\cite{eigen2014depth,ladicky2014pulling,liu2015learning}.  Several subsequent approaches~\cite{laina2016deeper,li2015depth,cao2016estimating} also treat depth estimation as a dense prediction problem and use popular fully-convolutional architectures such as FCN~\cite{long2015fullyconvolutional} or U-Net~\cite{ronneberger2015unet}.

Garg \ea~\cite{garg2016unsupervised} propose to use a calibrated stereo camera pair setup in which depth is produced as an intermediate output and the supervision comes from reconstruction of one image in a stereo pair from the input of the other. Since the images on the stereo rig have a fixed and known transformation, the depth can be learned from that functional relationship (plus some regularization). Other novel learning approaches, that also need more than one image for depth estimation are~\cite{xie2016deep3d,ranftl2016dense,ladicky2014pulling,kar2017learning,zbontar2016stereo}.  

Godard \ea~\cite{godard2017monodepth} offer an approach to learn single-view depth estimation using rectified stereo input during training. The disparity matching problem in a rectified stereo pair requires only a one-dimensional search.  The work by Ummenhofer \ea~\cite{ummenhofer2017demon} called DeMoN also addresses learning of depth from stereo data. Their method produces high-quality depth estimates from two unconstrained frames as input.  This work uses various forms of supervision including depth and optical flow.

Zhou \ea~\cite{zhou2017unsupervised} propose a novel approach for unsupervised learning of depth and ego-motion using only monocular video.  This setup is most aligned with our work as we similarly learn depth and ego-motion from monocular video in an unsupervised setting. Vijayanarasimhan \ea~\cite{vijayanarasimhan2017sfmnet} use a similar approach which additionally tries to learn the motion of a handful of objects in the scene.  Their work also allows for optional supervision by ground-truth depth or optical flow to improve performance.

Our work differs in taking the training process to three dimensions.  We present differentiable 3D loss functions which can establish consistency between the geometry of adjacent frames, and thereby improve depth and ego-motion estimation.

\section{Method}
\label{sec:method}

Our method learns depth and ego-motion from monocular video without supervision.  \reffig{approach} illustrates its different components.  At the core of our approach there is a novel loss function which is based on aligning the 3D geometry (point clouds) generated from adjacent frames (\refsec{3D}). Unlike 2D losses that enforce local photometric consistency, the 3D loss considers the entire scene and its geometry. We show how to efficiently backpropagate through this loss.

This section starts with discussing the geometry of the problem and how it is used to obtain differentiable losses.  It then describes each individual loss term.

\subsection{Problem Geometry}
\label{sec:problem}

At training time, the goal is to learn depth and ego-motion from a single monocular video stream. This problem can be formalized as follows: Given a pair of consecutive frames $X_{t-1}$ and $X_t$, estimate depth $D_{t-1}$ at time $t-1$, depth $D_{t}$ at time $t$, and the ego-motion $T_{t}$ representing the camera's movement (position and orientation) from time $t-1$ to $t$.

Once a depth estimate $D_t$ is available, it can be projected into a point cloud $Q_t$.  More specifically, the image pixel at coordinates $(i,j)$ with estimated depth $D_t^{ij}$ can be projected into a structured 3D point cloud
\begin{equation}\label{eq:qt}
Q_t^{ij} = D_t^{ij} \cdot K^{-1} [i, j, 1]^T,
\end{equation}
where $K$ is the camera intrinsic matrix, and the coordinates are homogeneous.

Given an estimate for $T_t$, the camera's movement from $t-1$ to $t$, $Q_t$ can be transformed to get an estimate for the previous frame's point cloud: $\hat{Q}_{t-1} = T_t Q_t$.  Note that the transformation applied to the point cloud is the inverse of the camera movement from $t$ to $t-1$.  $\hat{Q}_{t-1}$ can then be projected onto the camera at frame $t-1$ as $K \hat{Q}_{t-1}$.  Combining this transformation and projection with \refeq{qt} establishes a mapping from image coordinates at time $t$ to image coordinates at time $t-1$.  This mapping allows us to reconstruct frame $\hat{X}_t$ by warping $X_{t-1}$ based on $D_t, T_t$:
\begin{equation}\label{eq:r}
\hat{X}_t^{ij} = X_{t-1}^{\hat{i}\hat{j}}, [\hat{i}, \hat{j}, 1]^T = K T_t \big( D_t^{ij} \cdot K^{-1} [i, j, 1]^T \big)
\end{equation}
Following the approach in~\cite{zhou2017unsupervised,jaderberg2015spatial}, we compute $\hat{X}_t^{ij}$ by performing a soft sampling from the four pixels in $X_{t-1}$ whose coordinates overlap with $(\hat{i}, \hat{j})$.

This process is repeated in the other direction to project $D_{t-1}$ into a point cloud $Q_{t-1}$, and reconstruct frame $\hat{X}_{t-1}$ by warping $X_t$ based on $D_{t-1}$ and $T_t^{-1}$.

\subsection{Principled Masks}\label{sec:masks}

Computing $\hat{X}_{t}$ involves creating a mapping from image coordinates in $X_{t}$ to $X_{t-1}$.  However, due to the camera's motion, some pixel coordinates in $X_{t}$ may be mapped to coordinates that are outside the image boundaries in $X_{t-1}$.  With forward ego-motion, this problem is usually more pronounced when computing $\hat{X}_{t-1}$ from $X_t$.  Our experiments show that including such pixels in the loss degrades performance. Previous approaches have either ignored this problem, or tried to tackle it by adding a general-purpose mask to the network~\cite{garg2016unsupervised,zhou2017unsupervised,vijayanarasimhan2017sfmnet}, which is expected to exclude regions that are unexplainable due to any reason.  However, this approach does not seem to be effective and often results in edge artifacts in depth images (see \refsec{experiments}).

\begin{figure}[t]
  \begin{center}
	  \includegraphics[width=1.0\linewidth]{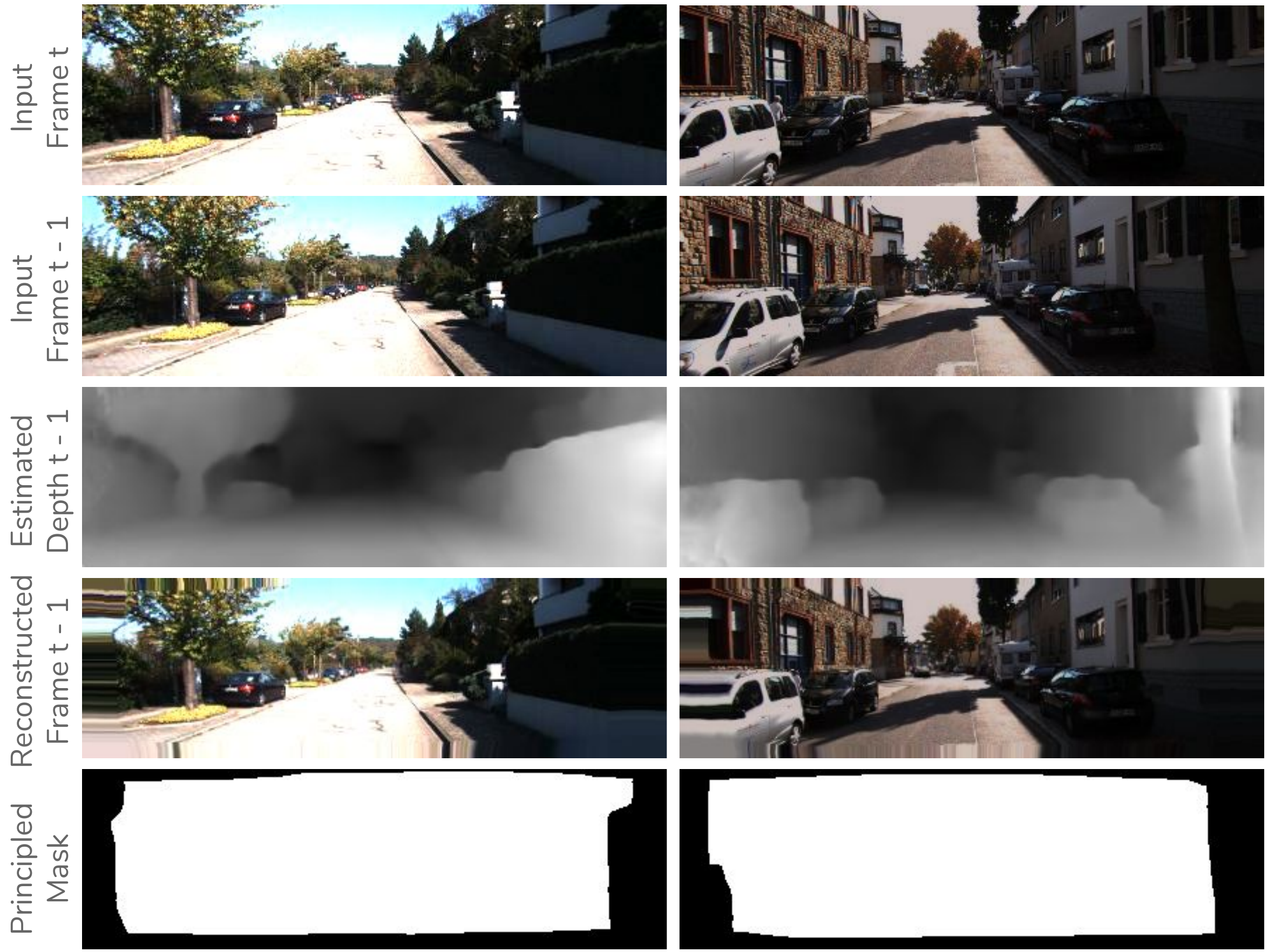}
  \end{center}
  \caption{Principled Masks.  The masks shown are examples of $M_{t-1}$, which indicate which pixel coordinates are valid when reconstructing $\hat{X}_{t-1}$ from $X_{t}$.  There is a complementary set of masks $M_{t}$ (not shown), which indicate the valid pixel coordinates when reconstructing $\hat{X}_{t}$ from $X_{t-1}$.}
  \label{fig:principled_masks}
\end{figure}

As \reffig{principled_masks} demonstrates, validity masks can be computed analytically from depth and ego-motion estimates.  For every pair of frames $X_{t-1}$, $X_t$, one can create a pair of masks $M_{t-1}$, $M_t$, which indicate the pixel coordinates where $\hat{X}_{t-1}$ and $\hat{X}_t$ are valid.

\subsection{Image Reconstruction Loss}

Comparing the reconstructed images $\hat{X}_{t}$, $\hat{X}_{t-1}$ to the input frames $X_{t}$, $X_{t-1}$ respectively produces a differentiable image reconstruction loss that is based on photometric consistency \cite{zhou2017unsupervised,garg2016unsupervised}, and needs to be minimized\footnote{\textbf{Note}: All losses mentioned in this section are repeated for times $t$ and $t-1$.  For brevity, we have left out the terms involving $t-1$.}:
\begin{equation}\label{eq:rec}
   L_\textup{rec} =  \sum_{ij} \|(X_{t}^{ij}-\hat{X}_{t}^{ij}) M_{t}^{ij}\|
\end{equation}

The main problem with this type of loss is that the process used to create $\hat{X}_{t}$ is an approximation---and, because differentiability is required, a relatively crude one.  This process is not able to account for effects such as lighting, shadows, translucence, or reflections. As a result, this loss is noisy and results in artifacts. Strong regularization is required to reduce the artifacts, which in turn leads to smoothed out predictions (see \refsec{experiments}). Learning to predict the adjacent frame directly would avoid this problem, but such techniques cannot generate depth and ego-motion predictions.

\subsection{A 3D Point Cloud Alignment Loss}
\label{sec:3D}

Instead of using $\hat{Q}_{t-1}$ or $\hat{Q}_{t}$ just to establish a mapping between coordinates of adjacent frames, we construct a loss function that directly compares point clouds $\hat{Q}_{t-1}$ to $Q_{t-1}$, or $\hat{Q}_{t}$ to $Q_{t}$.  This 3D loss uses a well-known rigid registration method, Iterative Closest Point (ICP)~\cite{chen1991object,besl1992amethod,rusinkiewicz2001efficient}, which computes a transformation that minimizes point-to-point distances between corresponding points in the two point clouds. 

ICP alternates between computing correspondences between two 3D point clouds (using a simple closest point heuristic), and computing a best-fit transformation between the two point clouds, given the correspondence. The next iteration then recomputes the correspondence with the previous iteration's transformation applied. Our loss function uses both the computed transformation and the final residual registration error after ICP's minimization.

Because of the combinatorial nature of the correspondence computation, ICP is not differentiable. As shown below, we can approximate its gradients using the products it computes as part of the algorithm, allowing us to backpropagate errors for both the ego-motion and depth estimates.

ICP takes as input two point clouds $A$ and $B$ (\eg $\hat{Q}_{t-1}$ and $Q_{t-1}$). Its main output is a best-fit transformation $T'$ which minimizes the distance between the transformed points in $A$ and their corresponding points in $B$: 
\begin{equation}\label{eq:icp}
  \argmin_{T'} \frac{1}{2} \sum_{ij} \|T'\cdot A^{ij} - B^{c(ij)}\|^2
\end{equation}
where $c(\cdot)$ denotes the point to point correspondence found by ICP. The secondary output of ICP is the residual $r^{ij} = A^{ij} - T'^{-1}\cdot B^{c(ij)}$, which reflects the residual distances between corresponding points after ICP's distance minimizing transform has been applied~\footnote{While we describe a point-to-point distance, we use the more powerful point-to-plane distance~\cite{chen1991object} as in the Point Cloud Library~\cite{rusu20113d}. The definition of the residual changes to include the gradient of the distance metric used, but it is still the gradient of the error.}.

\begin{figure}[t]
  \begin{center}
    \includegraphics[width=1.0\linewidth]{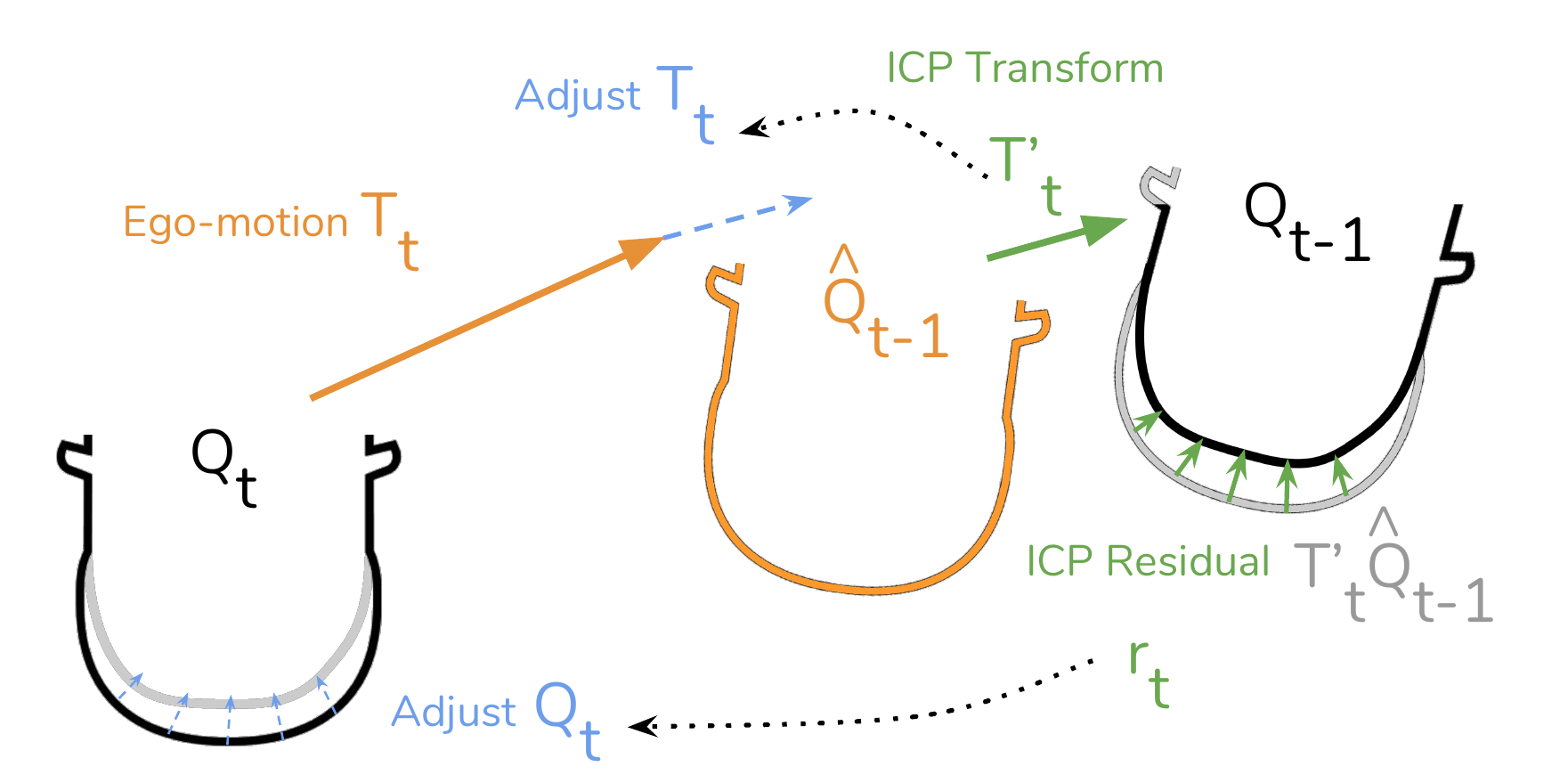}
  \end{center}
  \caption{The point cloud matching process and approximate gradients.  The illustration shows the top view of a car front with side mirrors. Given the estimated depth $D_{t}$ for timestep $t$ a point cloud $Q_{t}$ is created. This is transformed by the estimated ego-motion $T_t$ into a prediction the previous frame's point cloud, $\hat{Q}_{t-1}$. If ICP can find a better registration between $Q_{t-1}$ and $\hat{Q}_{t-1}$, we adjust our ego-motion estimate using this correction $T'_t$. Any residuals $r_t$ after registration point to errors in the depth map $D_{t}$, which are minimized by including $\| r_t \|_1$ in the loss.}
  \label{fig:icp_1d}
\end{figure}

\reffig{icp_1d} demonstrates how ICP is used in our method to penalize errors in the estimated ego-motion $T_t$ and depth $D_t$.  If the estimates $T_t$ and $D_t$ from the neural network are perfect, $\hat{Q}_{t-1}$ would align perfectly with $Q_{t-1}$.  When this is not the case, aligning $\hat{Q}_{t-1}$ to $Q_{t-1}$ with ICP produces a transform $T'_t$ and residuals $r_t$ which can be used to adjust $T_t$ and $D_t$ toward a better initial alignment.  More specifically, we use $T'_t$ as an approximation to the negative gradient of the loss with respect to the ego-motion $T_t$\footnote{Technically, $T'$ is not the negative gradient: it points in the direction of the minimum found by ICP, and not in the direction of steepest descent. Arguably, this makes it better than a gradient.}.  To correct the depth map $D_{t}$, we note that even after the correction $T'_t$ has been applied, moving the points in the direction $r_t$ would decrease the loss.  Of the factors that generate the points in $Q_{t}$ and thereby $\hat{Q}_{t-1}$, we can only change $D_{t}$.  We therefore use $r_t$ as an approximation to the negative gradient of the loss with respect to the depth $D_{t}$.  Note that this approximation of the gradient ignores the impact of depth errors on ego-motion and vice versa.  However, ignoring these second order effects works well in practice.  The complete 3D loss is then
\begin{equation}
L_\textup{3D} = \| T'_t - I \|_1 + \| r_t \|_1,
\label{eq:icp_loss}
\end{equation}
where $\|\cdot\|_1$ denotes the L1-norm, $I$ is the identity matrix.

\subsection{Additional Image-Based Losses}

Structured similarity (SSIM) is a commonly-used metric for evaluating the quality of image predictions. Similar to \cite{godard2017monodepth,zhao2015l2}, we use it as a loss term in the training process. It measures the similarity between two images patches $x$ and $y$ and is defined as $\textup{SSIM}(x,y)=\frac{(2\mu_x\mu_y+c_1)(2\sigma_{xy}+c_2)}{ (\mu_x^2+\mu_y^2+c_1)(\sigma_x+\sigma_y+c_2)}$, where $\mu_x,\sigma_x$ are the local means and variances~\cite{wang2004ssim}. In our implementation, $\mu$ and $\sigma$ are computed by simple (fixed) pooling, and $c_1=0.01^2$ and $c_2=0.03^2$. Since SSIM is upper bounded to one and needs to be maximized, we instead minimize
\begin{equation}
L_\textup{SSIM} = \sum_{ij} \big[ 1 - \textup{SSIM}(\hat{X}_t^{ij}, X_t^{ij}) \big] M_t^{ij}.
\label{eq:ssim}
\end{equation}

A depth smoothness loss is also employed to regularize the depth estimates. Similarly to~\cite{heise2013pm,godard2017monodepth} we use a depth gradient smoothness loss that takes into account the gradients of the corresponding input image:
\begin{equation}
L_\textup{sm} = \sum_{i,j} \| \partial_x D^{ij} \|e^{-\|\partial_x X^{ij}\|}+\| \partial_y D^{ij} \|e^{-\|\partial_y X^{ij}\|}
\label{eq:smooth}
\end{equation}

By considering the gradients of the image, this loss function allows for sharp changes in depth at pixel coordinates where there are sharp changes in the image. This is a refinement of the depth smoothness losses used by Zhou \ea~\cite{zhou2017unsupervised}.

\subsection{Learning Setup}

All loss functions are applied at four different scales $s$, ranging from the model's input resolution, to an image that is $\tfrac{1}{8}$ in width and height.  The complete loss is defined as:
\begin{equation}
L = \sum_{s} \alpha L^{s}_\textup{rec} + \beta L^s_\textup{3D} + \gamma L^s_\textup{sm} + \omega L^s_\textup{SSIM}
\label{eq:complete_loss}
\end{equation}
where $\alpha, \beta, \gamma, \omega$ are hyper-parameters, which we set to $\alpha = 0.85$, $\beta = 0.1$, $\gamma = 0.05$, and $\omega = 0.15$.

We adopt the SfMLearner architecture~\cite{zhou2017unsupervised}, which is in turn based on DispNet~\cite{mayer2016large}.  The neural network consists of two disconnected towers: A depth tower receives a single image with resolution $128 \times 416$ as input and produces a dense depth estimate mapping each pixel of the input to a depth value.  An ego-motion tower receives a stack of video frames as input, and produces an ego-motion estimate---represented by six numbers corresponding to relative 3D rotation and translation---between every two adjacent frames.  Both towers are fully convolutional.

At training time, a stack of video frames is fed to the model.  Following~\cite{zhou2017unsupervised}, in our experiments we use 3-frame training sequences, where our losses are applied over pairs of adjacent frames.  Unlike prior work, our 3D loss requires depth estimates from all frames.  However, at test time, the depth tower can produce a depth estimate from an individual video frame, while the ego-motion tower can produce ego-motion estimates from a stack of frames.

We use TensorFlow~\cite{abadi2016tensorflow} and the Adam optimizer with $\beta_1 = 0.9$, $\beta_2 = 0.999$, and $\alpha = 0.0002$.  In all experiments, models are trained for 20 epochs and checkpoints are saved at the end of each epoch.  The checkpoint which performs best on the validation set is then evaluated on the test set.

\begin{table*}[t]
  \centering
  \resizebox{0.9\textwidth}{!}{
  \begin{tabular}{|l|c|c|c||c|c|c|c|c|c|c|}
  \hline
  Method & Supervision & Dataset & Cap & \cellcolor{col1}Abs Rel & \cellcolor{col1}Sq Rel & \cellcolor{col1}RMSE  & \cellcolor{col1}RMSE log & \cellcolor{col2}$\delta < 1.25 $ & \cellcolor{col2}$\delta < 1.25^{2}$ & \cellcolor{col2}$\delta < 1.25^{3}$\\
  \hline 
  Train set mean & - & K & 80m & 0.361 & 4.826 & 8.102 & 0.377 & 0.638 & 0.804 & 0.894\\
  \hline 
  Eigen \ea\cite{eigen2014depth} Coarse & Depth & K & 80m & 0.214 & 1.605 & 6.563 & 0.292 & 0.673 & 0.884 & 0.957\\ 
  Eigen \ea\cite{eigen2014depth} Fine & Depth & K & 80m & 0.203 & 1.548 & 6.307 & 0.282 & 0.702 & 0.890 & 0.958\\ 
  Liu \ea\cite{liu2015learning} & Depth & K & 80m & 0.201 & 1.584 & 6.471 & 0.273 & 0.68 & 0.898 & 0.967\\
  \hline
  Zhou \ea\cite{zhou2017unsupervised} & - & K & 80m & 0.208 & 1.768 & 6.856 & 0.283 & 0.678 & 0.885 & 0.957 \\
  Zhou \ea\cite{zhou2017unsupervised} & - & CS + K & 80m & 0.198 & 1.836 & 6.565 & 0.275 & 0.718 & 0.901 & 0.960 \\
  Ours & - & K & 80m & 0.163 & 1.240 & 6.220 & 0.250 & 0.762 & 0.916 & 0.968 \\ 
  Ours & - & CS + K & 80m & \textbf{0.159} & \textbf{1.231} & \textbf{5.912} & \textbf{0.243} & \textbf{0.784} & \textbf{0.923} & \textbf{0.970} \\
  \hline
  Garg \ea\cite{garg2016unsupervised} & Stereo & K & 50m & 0.169 & 1.080 & 5.104 & 0.273 & 0.740 & 0.904 & 0.962 \\ 
  Zhou \ea\cite{zhou2017unsupervised} & - & K & 50m & 0.201 & 1.391 & 5.181 & 0.264 & 0.696 & 0.900 & 0.966 \\
  Zhou \ea\cite{zhou2017unsupervised} & - & CS + K & 50m & 0.190 & 1.436 & 4.975 & 0.258 & 0.735 & 0.915 & 0.968 \\
  Ours & - & K & 50m & 0.155 & 0.927 & 4.549 & 0.231 & 0.781 & 0.931 & \textbf{0.975} \\ 
  Ours & - & CS + K & 50m & \textbf{0.151} & \textbf{0.949} & \textbf{4.383} & \textbf{0.227} & \textbf{0.802} & \textbf{0.935} & 0.974 \\
  \hline
  \end{tabular}
  }
  \vspace{10pt}
  \caption{Depth evaluation metrics over the KITTI Eigen~\cite{eigen2014depth} test set.  Under the Dataset column, K denotes training on KITTI ~\cite{geiger2012we} and CS denotes training on Cityscapes \cite{cordts2016cityscapes}.  $\delta$ denotes the ratio between estimates and ground truth.  All results, except \cite{eigen2014depth}, use the crop from \cite{garg2016unsupervised}.
}
  \label{tab:kitti_eigen}
  \vspace{-10pt}
\end{table*}

\begin{figure*}[t]
  \begin{center}
    \includegraphics[width=1.0\linewidth]{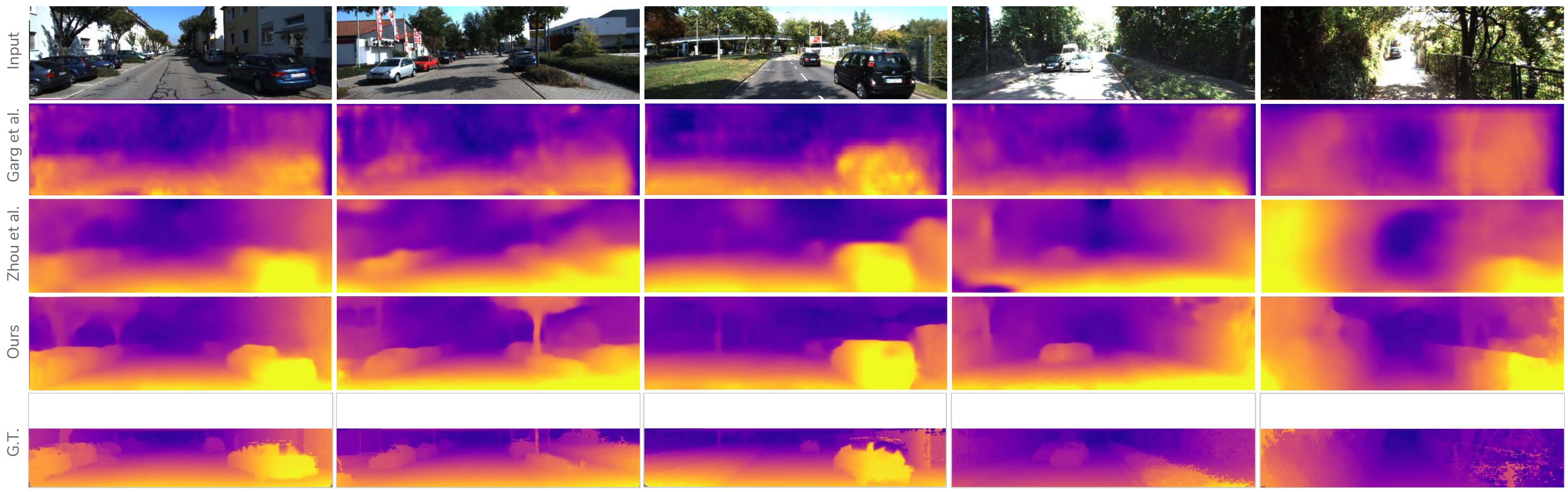}
  \end{center}
  \vspace{-10pt}
  \caption{Sample depth estimates from the KITTI Eigen test set, generated by our approach (4th row), compared to Garg \ea~\cite{garg2016unsupervised}, Zhou~\ea~\cite{zhou2017unsupervised}, and ground truth~\cite{geiger2013vision}. Best viewed in color.}
  \label{fig:grid}
\end{figure*}

\begin{figure}[t!]
  \begin{center}
    \includegraphics[width=1.0\linewidth]{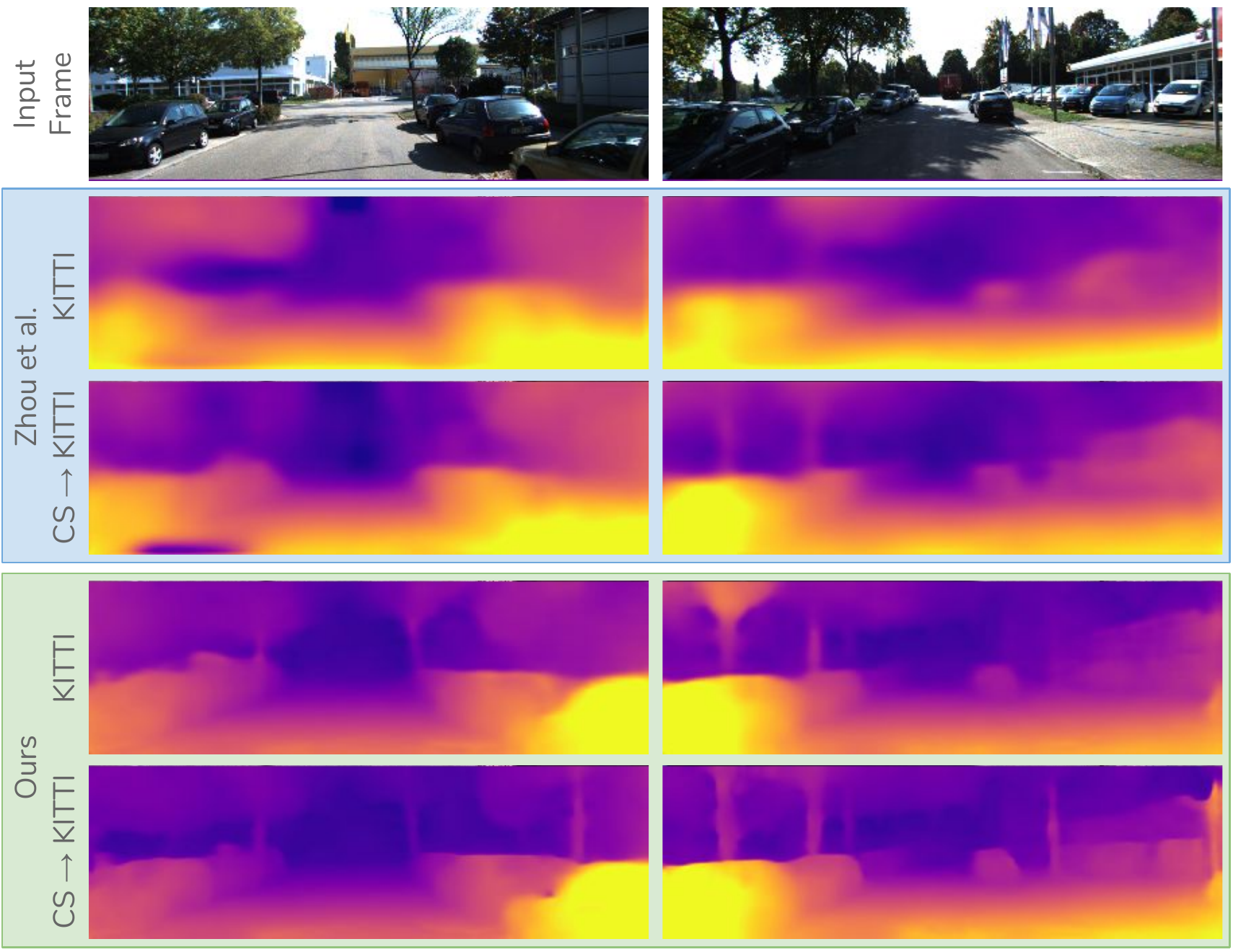}
  \end{center}
  \caption{Comparison of models trained only on KITTI vs. models pre-trained on Cityscapes and then fine-tuned on KITTI.  The first two rows show depth images produced by models from~\cite{zhou2017unsupervised}.  These images are generated by us using models trained by~\cite{zhou2017unsupervised}.  The bottom two rows show depth images produced by our method.}
  \label{fig:vs-sfmlearner}
\end{figure}

\section{Experiments}
\label{sec:experiments}

\subsection{Datasets}
\label{sec:datasets}

\textbf{KITTI}. We use the KITTI dataset~\cite{geiger2013vision} as the main training and evaluation dataset. This dataset is the most common benchmark used in prior work for evaluating depth and ego-motion accuracy~\cite{garg2016unsupervised,zhou2017unsupervised,godard2017monodepth,ummenhofer2017demon}.  The KITTI dataset includes a full suite of data sources such as stereo video, 3D point clouds from LIDAR, and the vehicle trajectory.  We use only a single (monocular) video stream for training.  The point clouds and vehicle poses are used only for evaluation of trained models.  We use the same training/validation/test split as \cite{zhou2017unsupervised}: about 40k frames for training, 4k for validation, and 697 test frames from the Eigen~\cite{eigen2014depth} split.

\textbf{Uncalibrated Bike Video Dataset}.  We created a new dataset by recording some videos using a hand-held phone camera while riding a bicycle.  This particular camera offers no stabilization. The videos were recorded at 30fps, with a resolution of 720$\times$1280.  Training sequences were created by selecting frames at 5fps to roughly match the motion in KITTI.  We used all $91,866$ frames from the videos without excluding any particular segments.  We constructed an intrinsic matrix for this dataset based on a Google search for ``iPhone 6 video horizontal field of view'' (50.9$^{\circ}$) and without accounting for lens distortion.  This dataset is available on the project website.

\subsection{Evaluation of Depth Estimation}
\label{sec:depth}

\reffig{grid} compares sample depth estimates produced by our trained model to other unsupervised learning methods, including the state-of-the-art results by ~\cite{zhou2017unsupervised}.

\reftab{kitti_eigen} quantitatively compares our depth estimation results against prior work (some of which use supervision).  The metrics are computed over the Eigen~\cite{eigen2014depth} test set.  The table reports separate results for a depth cap of 50m, as this is the only evaluation reported by Garg \ea~\cite{garg2016unsupervised}.  When trained only on the KITTI dataset, our model lowers the mean absolute relative depth prediction error from 0.208 \cite{zhou2017unsupervised} to 0.163, which is a significant improvement.  Furthermore, this result is close to the state-of-the-art result of 0.148 by Godard \ea~\cite{godard2017monodepth}, obtained by training on rectified stereo images with known camera baseline.

Since our primary baseline~\cite{zhou2017unsupervised} reports results for pre-training on Cityscapes~\cite{cordts2016cityscapes} and fine-tuning on KITTI, we replicate this experiment as well.  \reffig{vs-sfmlearner} shows the increase in quality of depth estimates as a result of pre-training on Cityscapes.  It also visually compares depth estimates from our models with the corresponding models by Zhou \ea~\cite{zhou2017unsupervised}.  As \reffig{vs-sfmlearner} and \reftab{kitti_eigen} show, our proposed method achieves significant improvements.  The mean inference time on an input image of size $128 \times 416$ is $10.5$ ms on a GeForce GTX 1080.

\subsection{Evaluation of the 3D Loss}

\begin{figure}
  \includegraphics[width=0.23\textwidth]{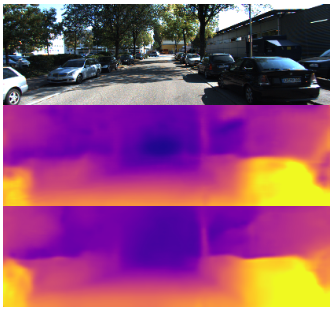}
  \includegraphics[width=0.23\textwidth]{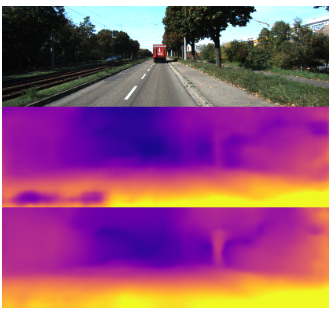}
  \caption{Example depth estimation results from training without the 3D loss (middle), and with the 3D loss (bottom).}
  \label{fig:icp_comp}
\end{figure}

\reffig{icp_comp} shows sample depth images produced by models which are trained with and without the 3D loss.  As the sample image shows, the additional temporal consistency enforced by the 3D loss can reduce artifacts in low-texture regions of the image.

\begin{figure}[t]
  \begin{center}
    \includegraphics[width=0.9\linewidth]{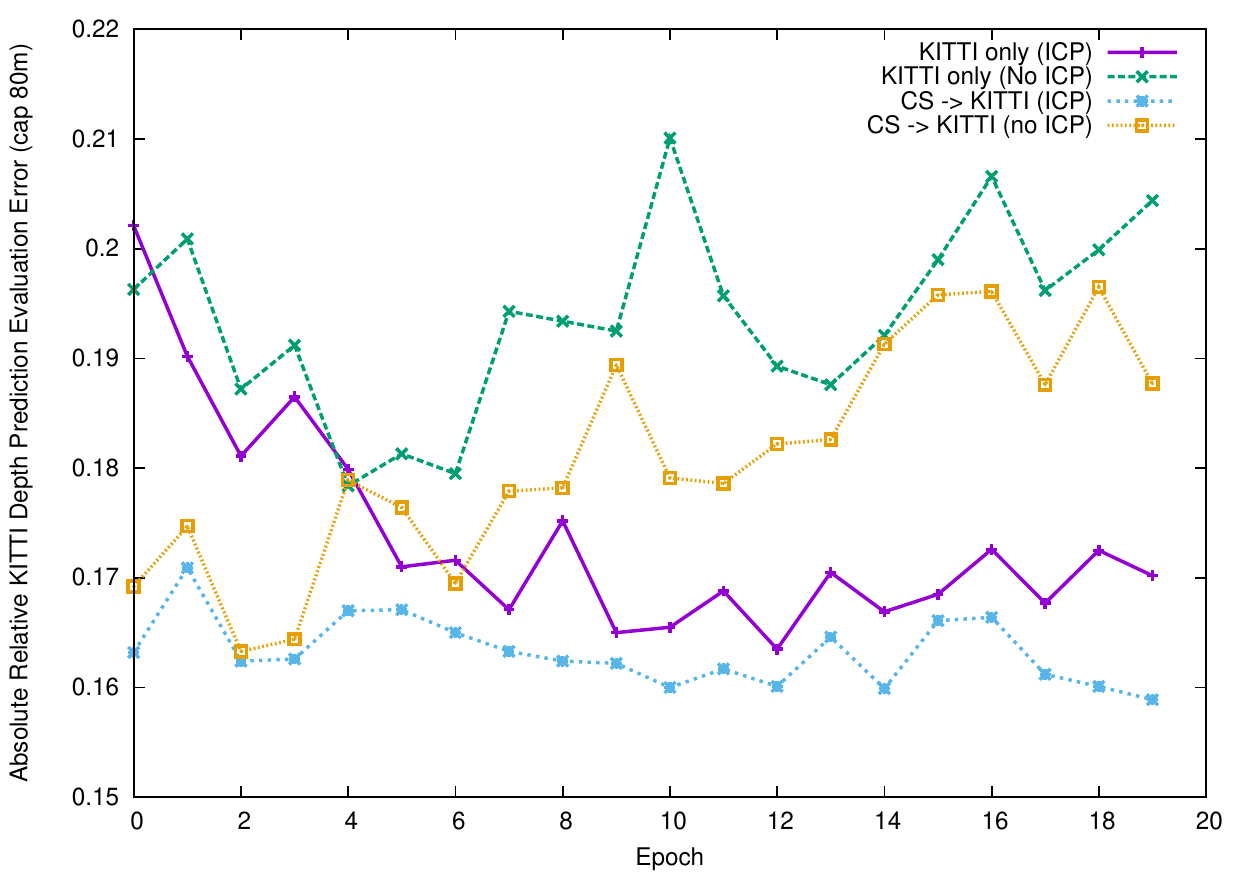}
  \end{center}
  \caption{Evolution of depth validation error over time when training our model with and without the 3D loss.  Training on KITTI and on Cityscapes + KITTI are shown.  Using the 3D loss lowers the error and also reduces overfitting.}
  \label{fig:cs-to-k-80}
\end{figure}

\reffig{cs-to-k-80} plots the validation error from each model over time as training progresses.  The points show depth error at the end of different training epochs on the validation set---and not the test set, which is reported in \reftab{kitti_eigen}. As the plot shows, using the 3D loss improves performance notably across all stages of training.  It also shows that the 3D loss has a regularizing effect, which reduces overfitting.  In contrast, just pre-training on the larger Cityscapes dataset is not sufficient to reduce overfitting or improve depth quality.

\begin{table}[t]
  \centering
  \scalebox{0.88}{
  \begin{tabular}{lcc}
  \toprule
  \textbf{Method} & \textbf{Seq. $09$} & \textbf{Seq. $10$}
  \tabularnewline
  \midrule
  \textbf{ORB-SLAM (full)} & $0.014 \pm 0.008$ & $\mathbf{0.012 \pm 0.011}$  \tabularnewline
  \midrule
  \textbf{ORB-SLAM (short)} & $0.064 \pm 0.141$ & $0.064 \pm 0.130$ \tabularnewline
  \textbf{Mean Odom.} & $0.032 \pm 0.026$ & $0.028 \pm 0.023$ \tabularnewline
  \textbf{Zhou \ea\cite{zhou2017unsupervised}} (5-frame) & $0.021 \pm 0.017$ & $0.020 \pm 0.015$ \tabularnewline
  \textbf{Ours, no ICP} (3-frame) & $0.014 \pm 0.010$ & $0.013 \pm 0.011$ \tabularnewline
  \textbf{Ours, with ICP} (3-frame) & $\mathbf{0.013 \pm 0.010}$ & $\mathbf{0.012 \pm 0.011}$ \tabularnewline
  \bottomrule
  \end{tabular}}
  \vspace{0.1cm}
  \caption{Absolute Trajectory Error (ATE) on the KITTI odometry dataset averaged over all multi-frame snippets (lower is better). Our method significantly outperforms the baselines with the same input setting.  It also matches or outperforms ORB-SLAM (full) which uses strictly more data.}
  \label{tab:odom}
  \vspace{-10pt}
\end{table}

\begin{table*}[t]
  \centering
  \resizebox{0.8\textwidth}{!}{
  \begin{tabular}{|l|c|c| |c|c|c|c|c|c|c|c|}
  \hline
  Method & Dataset & Cap & \cellcolor{col1}Abs Rel & \cellcolor{col1}Sq Rel & \cellcolor{col1}RMSE  & \cellcolor{col1}RMSE log & \cellcolor{col2}$\delta < 1.25 $ & \cellcolor{col2}$\delta < 1.25^{2}$ & \cellcolor{col2}$\delta < 1.25^{3}$\\
  \hline
  All losses & CS + K & 80m & \textbf{0.159} & \textbf{1.231} & \textbf{5.912} & \textbf{0.243} & \textbf{0.784} & \textbf{0.923} & \textbf{0.970} \\
  \hline
  All losses & K & 80m & 0.163 & 1.240 & 6.220 & 0.250 & 0.762 & 0.916 & 0.968 \\ 
  No ICP loss & K & 80m & 0.175 & 1.617 & 6.267 & 0.252 & 0.759 & 0.917 & 0.967 \\ 
  No SSIM loss & K & 80m & 0.183 & 1.410 & 6.813 & 0.271 & 0.716 & 0.899 & 0.961 \\ 
  No Principled Masks & K & 80m & 0.176 & 1.386 & 6.529 & 0.263 & 0.740 & 0.907 & 0.963 \\ 
  \hline
  Zhou \ea\cite{zhou2017unsupervised}  & K & 80m & 0.208 & 1.768 & 6.856 & 0.283 & 0.678 & 0.885 & 0.957 \\
  Zhou \ea\cite{zhou2017unsupervised}  & CS + K & 80m & 0.198 & 1.836 & 6.565 & 0.275 & 0.718 & 0.901 & 0.960 \\
  \hline
  All losses & Bike & 80m & 0.211 & 1.771 & 7.741 & 0.309 & 0.652 & 0.862 & 0.942 \\ 
  No ICP loss & Bike & 80m & 0.226 & 2.525 & 7.750 & 0.305 & 0.666 & 0.871 & 0.946 \\ 
  \hline
  \end{tabular}
  }
  \vspace{10pt}
  \caption{Depth evaluation metrics over the KITTI Eigen~\cite{eigen2014depth} test set for various versions of our model.  Top: Our best model.  Middle: Ablation results where individual loss components are excluded. Bottom: Models trained only on the bike dataset.}
  \label{tab:kitti_eigen2}
  \vspace{-10pt}
\end{table*}

\subsection{Evaluation of Ego-Motion}

During the training process, depth and ego-motion are learned jointly and their accuracy is inter-dependent. ~\reftab{odom} reports the ego-motion accuracy of our models over two sample sequences from the KITTI odometry dataset.  Our proposed method significantly outperforms the unsupervised method by~\cite{zhou2017unsupervised}.  Moreover, it matches or outperforms the supervised method of ORB-SLAM, which uses the entire video sequence.

\subsection{Learning from Bike Videos}

\begin{figure}[t]
  \begin{center}
    \includegraphics[width=0.9\linewidth]{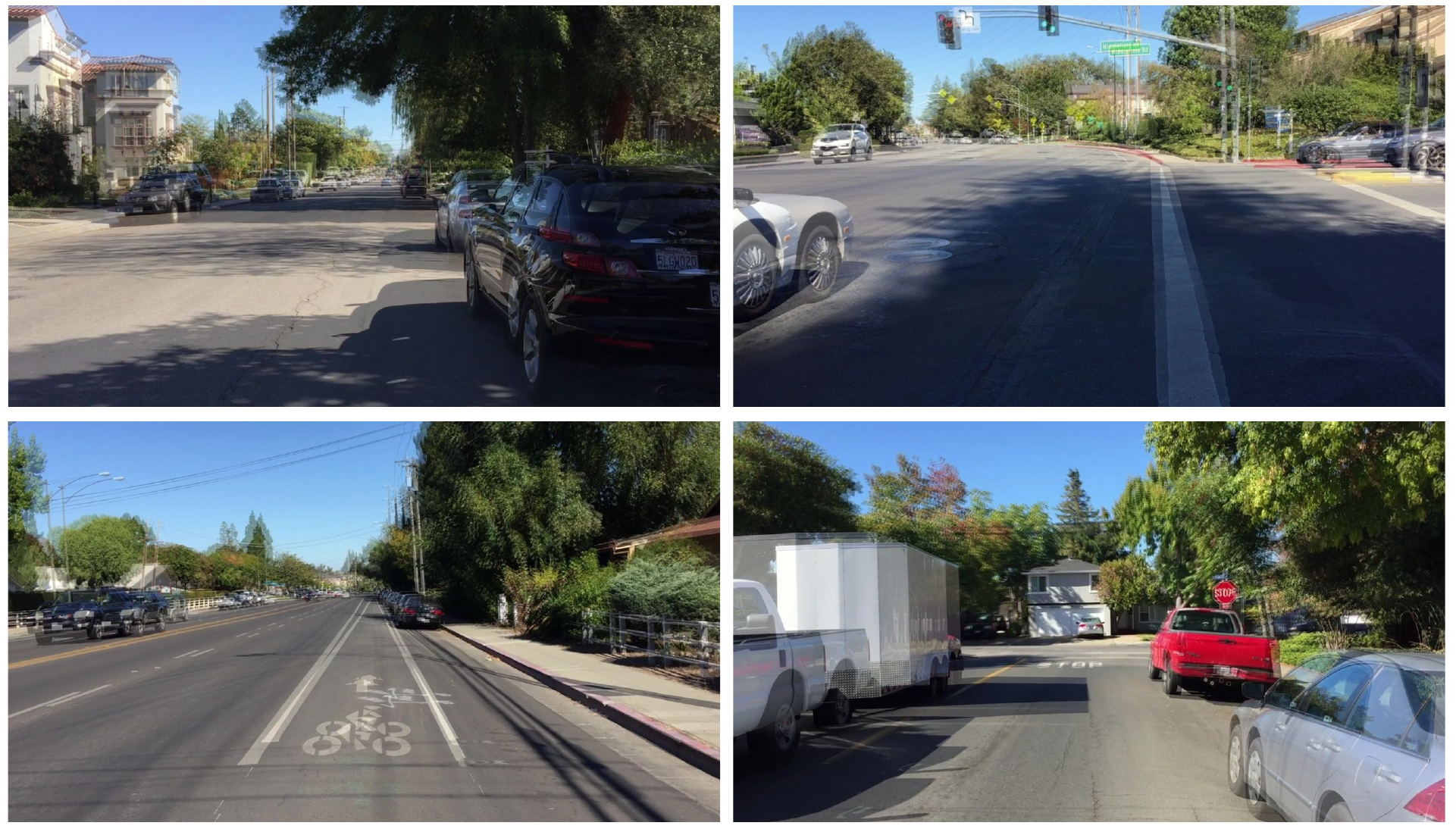}
  \end{center}
  \caption{Composite of two consecutive frames from the Bike dataset.  Since the phone is hand-held, the motion is less stable compared to existing driving datasets. Best viewed in color.}
  \label{fig:bike_frames}
\end{figure}

To demonstrate that our proposed method can use any video with ego-motion as training data, we recorded a number of videos using a \emph{hand-held} phone camera while riding a bicycle. \reffig{bike_frames} shows sample frames from this dataset.

\begin{figure}[t]
  \begin{center}
    \includegraphics[width=1.0\linewidth]{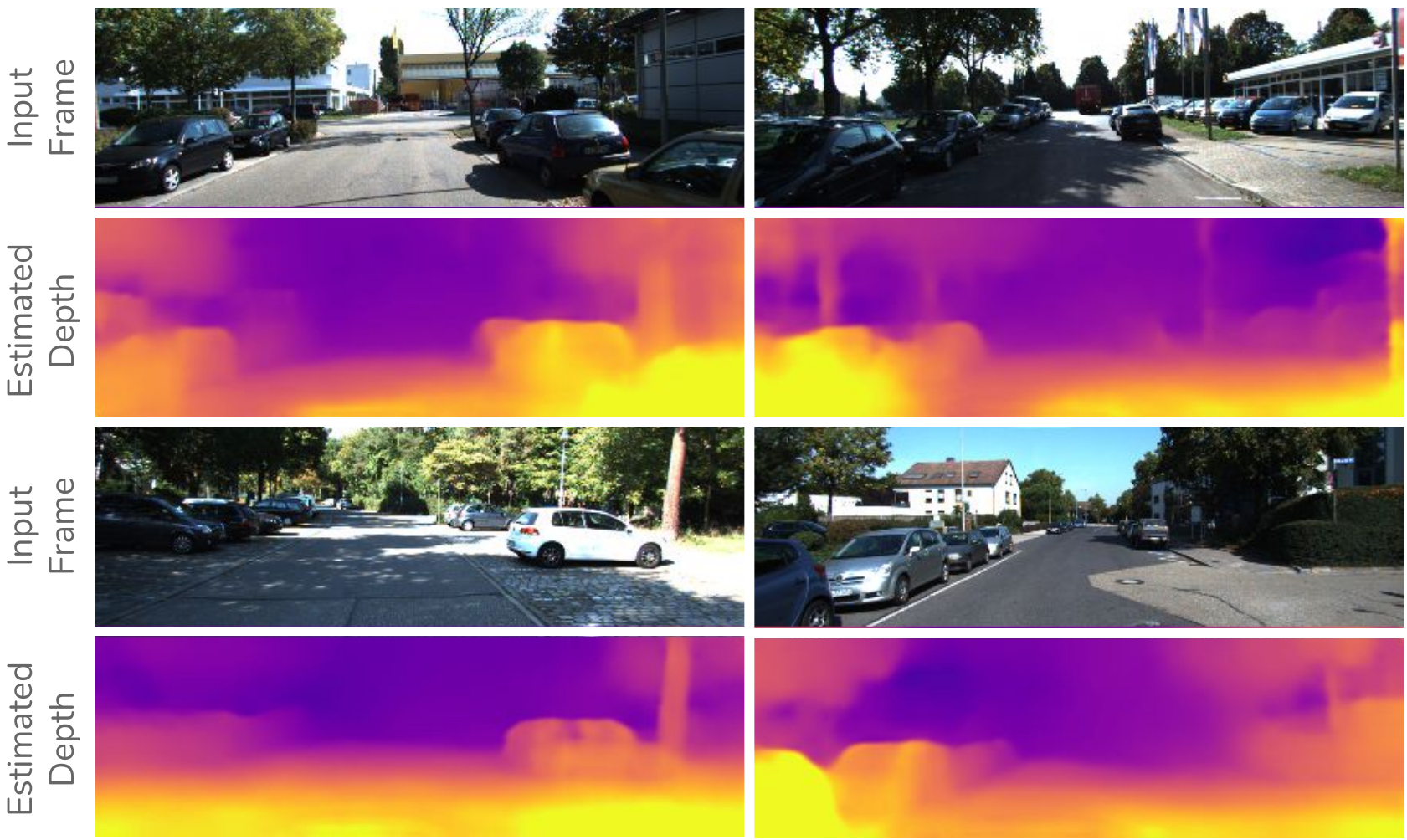}
  \end{center}
  \caption{Sample depth estimates produced from KITTI frames by the model trained only on the Bike video dataset.}
  \label{fig:bike}
\end{figure}

We trained our depth and ego-motion model only on the Bike videos.  We then evaluated the trained model on KITTI. Note that no fine-tuning is performed.  \reffig{bike} show sample depth estimates for KITTI frames produced by the model trained on Bike videos.  The Bike dataset is quite different from the KITTI dataset ($\sim 51^{\circ}$ vs. $\sim 81^{\circ}$ FOV, no distortion correction vs. fully rectified images, US vs. European architecture/street layout, hand-held camera vs. stable motion).  Yet, as \reftab{kitti_eigen2} and \reffig{bike} show, the model trained on Bike videos is close in quality to the best unsupervised model of~\cite{zhou2017unsupervised}, which is trained on KITTI itself.

\begin{figure}[t]
  \begin{center}
    \includegraphics[width=0.9\linewidth]{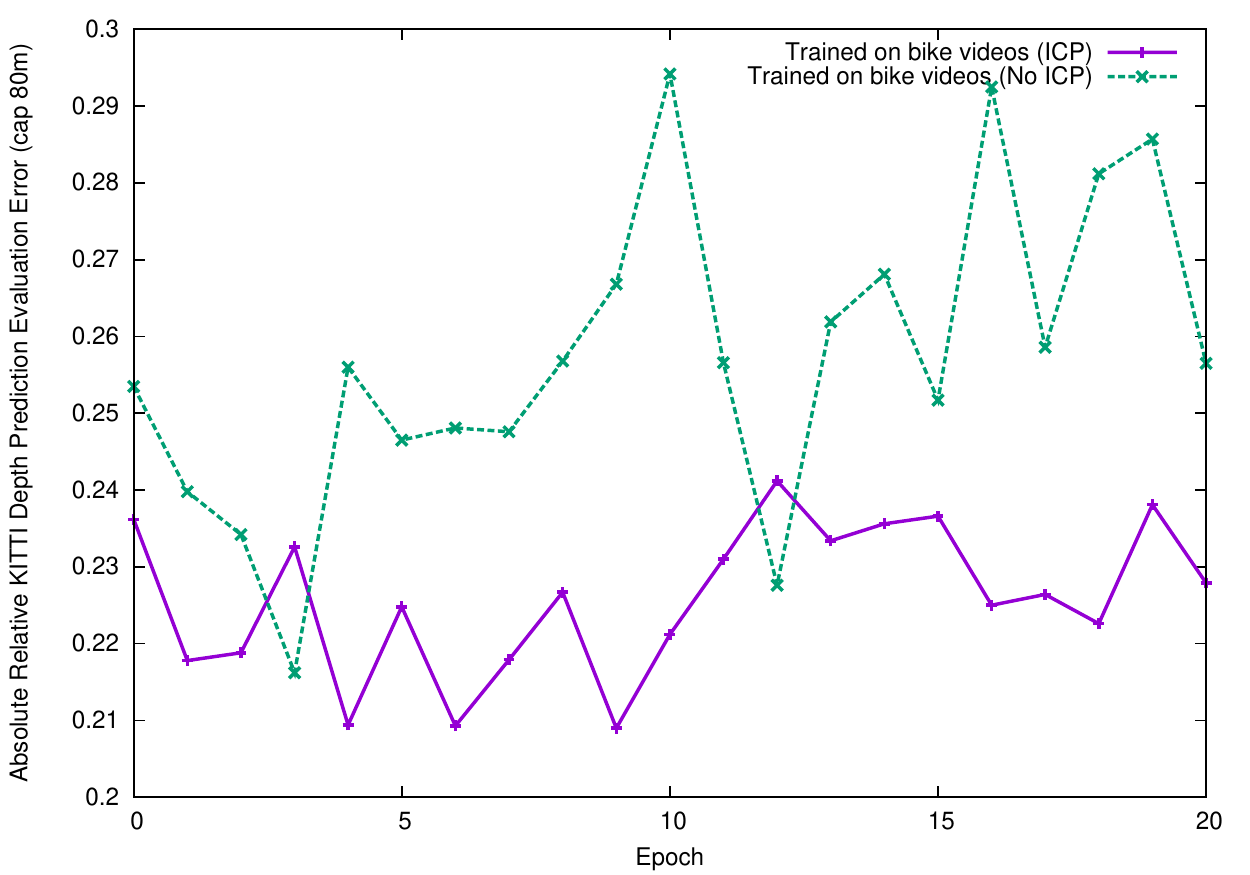}
  \end{center}
  \caption{Evolution of KITTI depth validation error for models trained only on the Bike Dataset, with and without the 3D loss.}
  \label{fig:bike-80}
\end{figure}

\reffig{bike-80} shows the \emph{KITTI validation error} for models trained on Bike videos.  It verifies that the 3D loss improves learning and reduces overfitting on this dataset as well.

\subsection{Ablation Experiments}

\begin{figure}[t]
  \begin{center}
    \includegraphics[width=0.9\linewidth]{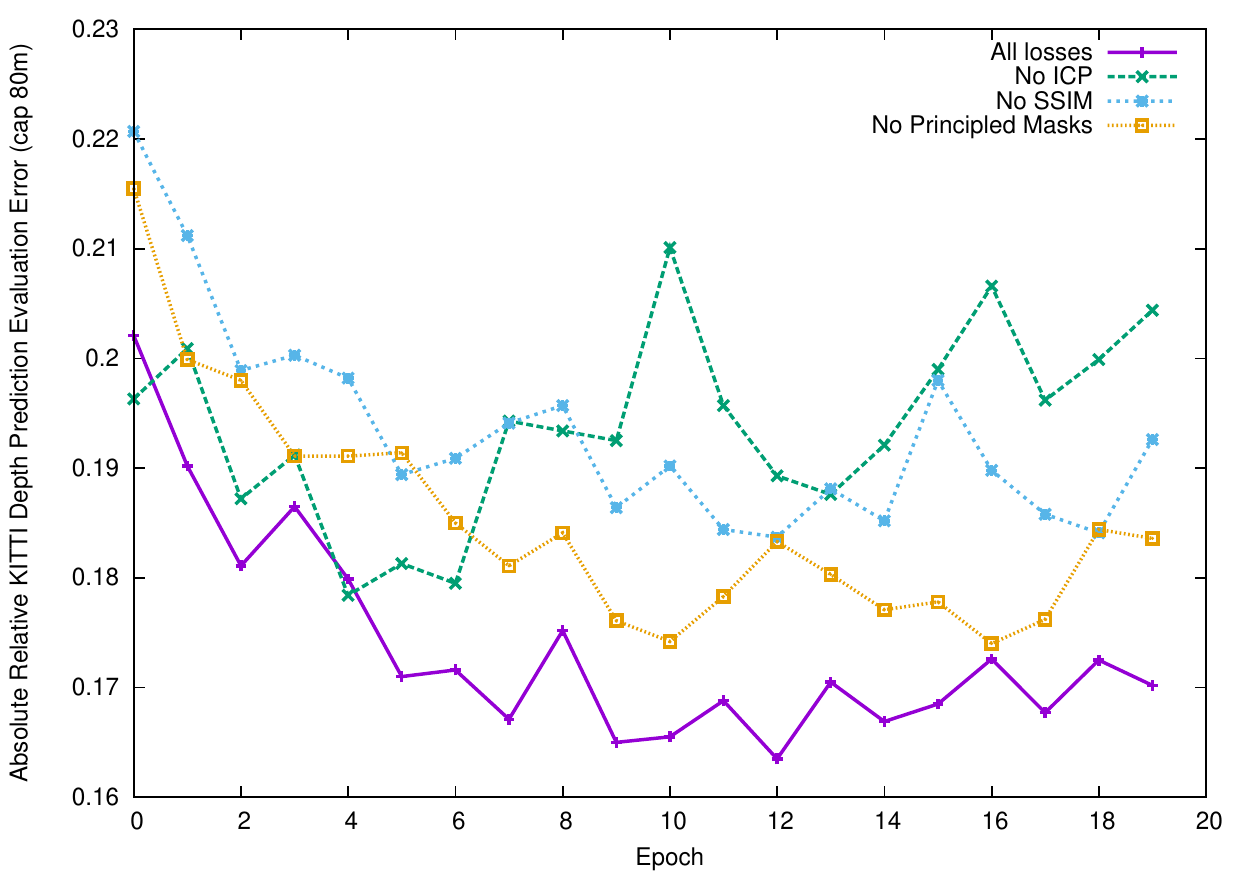}
  \end{center}
  \caption{KITTI depth validation error for ablation experiments comparing a model trained with all losses against models missing specific loss components.}
  \label{fig:ablation}
\end{figure}

In order to study the importance of each component in our method, we trained and evaluated a series of models, each missing one component of the loss function.  The experiment results in \reftab{kitti_eigen2} and \reffig{ablation} show that the 3D loss and SSIM components are essential.  They also show that removing the masks hurts the performance.  

\section{Conclusions and Future Work}
\label{sec:conclusion}
We proposed a novel unsupervised algorithm for learning depth and ego-motion from monocular video. Our main contribution is to explicitly take the 3D structure of the world into consideration. We do so using a novel loss function which can align 3D structures across different frames.  The proposed algorithm needs only a single monocular video stream for training, and can produce depth from a single image at test time.

The experiments on the Bike dataset demonstrate that our approach can be applied to learn depth and ego-motion from diverse datasets.  Because we require no rectification and our method is robust to lens distortions, lack of stabilization, and other features of low-end cameras, training data can be collected from a large variety of sources, such as public videos on the internet. 

If an object moves between two frames, our loss functions try to explain its movement by misestimating its depth.  This leads to learning biased depth estimates for that type of object.  Similar to prior work~\cite{zhou2017unsupervised}, our approach does not explicitly handle largely dynamic scenes.  Detecting and handling moving objects is our goal for future work.

Lastly, the principled masks can be extended to account for occlusions and disocclusions resulting from change of viewpoint between adjacent frames.

\section*{Acknowledgments}

We thank Tinghui Zhou and Cl\'ement Godard for sharing their code and results, the Google Brain team for discussions and support, and Parvin Taheri and Oscar Ezhdeha for their help with recording videos.

{\small
\bibliographystyle{ieee}
\bibliography{main}
}

\end{document}